\documentclass[conference]{IEEEtran}
\IEEEoverridecommandlockouts
\usepackage{cite}
\usepackage{amsmath,amssymb,amsfonts}
\usepackage{algorithmic}
\usepackage{graphicx}
\usepackage{textcomp}
\usepackage{xcolor}
\def\BibTeX{{\rm B\kern-.05em{\sc i\kern-.025em b}\kern-.08em
    T\kern-.1667em\lower.7ex\hbox{E}\kern-.125emX}}
    
\usepackage{epstopdf}
\usepackage{url}
\usepackage{algorithm}
\usepackage{bm}
\usepackage{multirow}
\usepackage{ragged2e}

\begin{document}

\title{INVITED:\\New Directions in Distributed Deep Learning: Bringing the Network at Forefront of IoT Design\thanks{This preprint is for personal use only. The official article will appear in proceedings of Design Automation Conference (DAC), 2020. This work was presented at the DAC 2020 special session on Edge-to-Cloud Neural Networks for Machine Learning Applications in Future IoT Systems.}}

\author{\IEEEauthorblockN{Kartikeya Bhardwaj}
\IEEEauthorblockA{
\textit{Arm Inc.}\\
San Jose, CA, USA \\
kartikeya.bhardwaj@arm.com}
\and
\IEEEauthorblockN{Wei Chen}
\IEEEauthorblockA{
\textit{Carnegie Mellon University}\\
Pittsburgh, PA, USA \\
weic3@andrew.cmu.edu}
\and
\IEEEauthorblockN{Radu Marculescu}
\IEEEauthorblockA{
\textit{The University of Texas at Austin}\\
Austin, TX, USA \\
radum@utexas.edu}
}

\maketitle

\begin{abstract}
In this paper, we first highlight three major challenges to large-scale adoption of deep learning at the edge: (\textit{i})~Hardware-constrained IoT devices, (\textit{ii})~Data security and privacy in the IoT era, and (\textit{iii})~Lack of \textit{network-aware} deep learning algorithms for distributed inference across multiple IoT devices. We then provide a unified view targeting three research directions that naturally emerge from the above challenges: (1)~Federated learning for training deep networks, (2)~Data-independent deployment of learning algorithms, and (3)~Communication-aware distributed inference. We believe that the above research directions need a network-centric approach to enable the edge intelligence and, therefore, fully exploit the true potential of IoT.
\end{abstract}


\section{Introduction}\label{sec:intro}
An estimated one trillion Internet-of-Things (IoT) devices are expected to impact several market segments by 2035~\cite{arm}.  
Such a rapid growth in IoT devices necessitates new breakthroughs in Machine Learning (ML) and Artificial Intelligence (AI) in order to fully exploit the compute power offered by a trillion devices. Specifically, there are three major challenges to rapid adoption of deep learning at the edge:
\begin{itemize}
    \item \textbf{Hardware-constrained IoT devices:} Typically, IoT devices are memory-limited (\textit{e.g.}, only a few hundred KB memory is available) and run at low operating frequencies for high energy efficiency. Since challenges like hardware constraints have been surveyed in~\cite{edgComp}, our focus here is on the new research directions from a deep learning perspective.
    \item \textbf{Data security and privacy:} With data at the edge becoming increasingly personal, it is now extremely important for the AI utilizing this data to operate \textit{locally} on user devices. Therefore, instead of sending the private data to the cloud, new techniques are needed to enable both on-device training and inference of deep learning models without compromising data privacy.
    \item \textbf{\textit{Network-aware} deep learning:}
    By very definition, IoT refers to a \textit{network} of devices. Yet, most of the techniques proposed to date, especially for inference, focus on reducing the computational requirements of deep neural networks for a \textit{single} device and do \textit{not} take the network  into account. Hence, new research that can effectively exploit the benefits of the network is crucial. 
\end{itemize}

From a deep learning standpoint, \textit{model compression} aims to reduce the size of deep networks in order to meet the hardware constraints of IoT devices without sacrificing accuracy~\cite{deepComp, hintonKD, rudy}. However, in practice, additional challenges like data privacy and network considerations often severely limit the widespread deployment of deep learning on IoT devices~\cite{bhardwaj2019edgeai}. More specifically, the above fundamental challenges manifest themselves in training and inference, which are both integral parts of any learning-based system. We next explain how these challenges are closely intertwined with the emerging directions in training and inference of deep neural networks at the edge.\vspace{1mm}

\noindent
\textbf{Data privacy and network of devices in regards to training:}
To prevent sending private data to the cloud and exploit the network of devices, a new distributed training paradigm called Federated Learning (FL)~\cite{konevcny2016federated,konevcny2015federated} has emerged recently. FL first trains ML models on-device using the local data. Then, the model (instead of private data) is sent to the cloud for a global update; this global model is subsequently sent back to the edge devices. There are several challenges in this problem space such as (\textit{i}) \textit{statistical heterogeneity} that happens when data is not independently and identically distributed (non-IID) across users, and (\textit{ii}) \textit{systems heterogeneity} when the devices that are training have different computational and communication capabilities. Therefore, FL aims to alleviate the data privacy issues while exploiting a vast network of devices to train the models. A good review of these problems is given in~\cite{li2019federated}.\vspace{1mm} 
\begin{figure*}[!t]
\centering
\includegraphics[width=\textwidth]{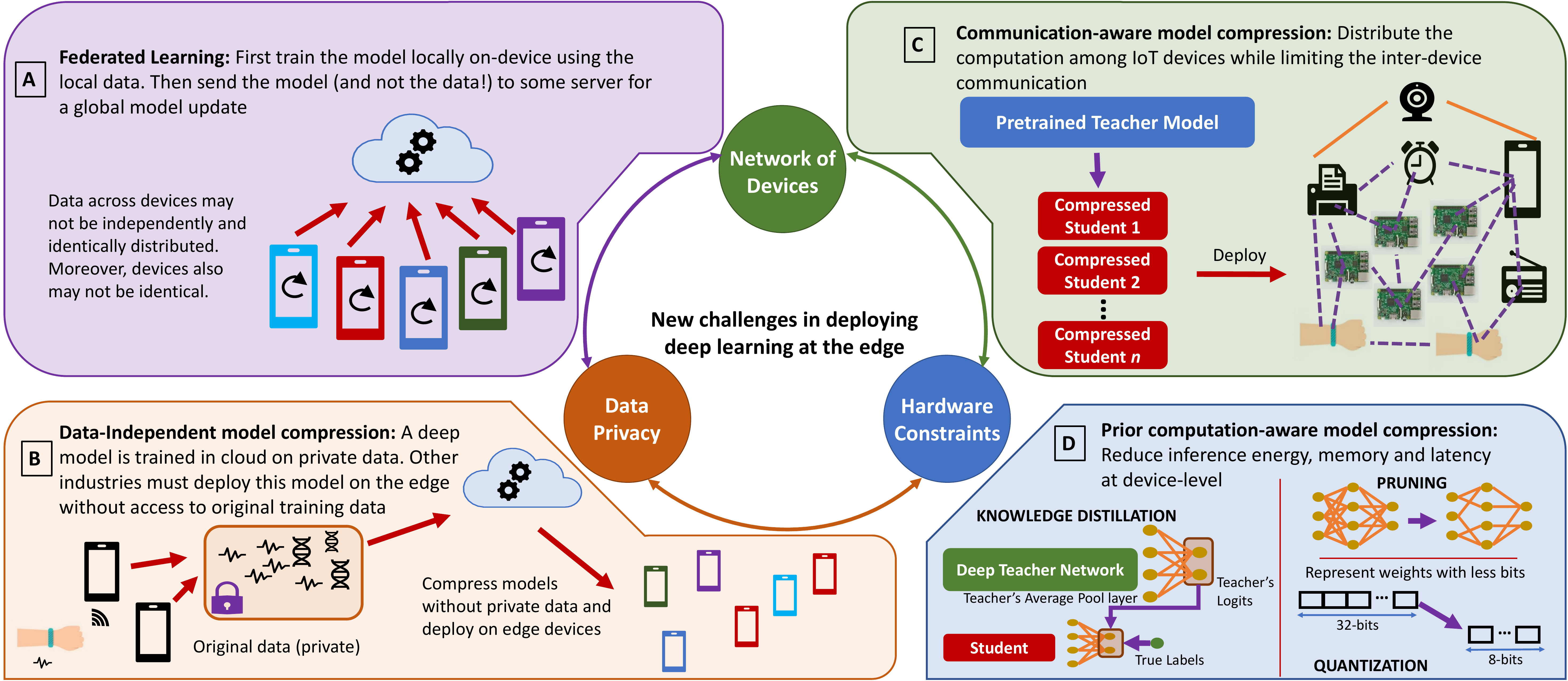}\vspace{-2mm}
	\caption{Bringing the network at the forefront of IoT design. Major challenges in deploying deep learning at the edge include the lack of network-aware algorithms, data privacy, and hardware constraints of IoT devices. (a) Federated Learning is a new distributed training paradigm at the intersection of data privacy and network of devices, (b) Data-independent model compression aims to compress deep networks without using private datasets -- This inference problem is at the intersection of hardware constraints and data privacy, (c) Communication-aware model compression is a new distributed inference paradigm at the intersection of hardware constraints and network of devices: Exploit the network of devices to collaboratively obtain low-latency, high-accuracy intelligence at the edge, (d) Prior model compression methods focus on the hardware constraints at device-level, but not the other two critical challenges: Key techniques include Pruning, Quantization, and Knowledge Distillation (KD).}
\label{flow}
\end{figure*}

\noindent
\textbf{Hardware constraints and data privacy in regards to inference:} 
It might appear that for resource-constrained inference, data privacy is not a big challenge as deep learning inference, by definition, means that we are not sending the data to the cloud. Indeed, deploying deep neural networks at the edge requires model compression. However, most model compression techniques rely on access to the original training dataset or some alternate data; for many applications (\textit{e.g.}, image processing on medical images, speech recognition, \textit{etc.}), such datasets are private. 

Complex deep learning models can result from training on huge private datasets. Hence, the industries deploying such models at the edge must compress them without accessing the original training dataset or any alternate dataset\footnote{It is possible for the industry deploying a model at the edge to collect alternate datasets for model compression. However, this may not always be possible, or can be very time consuming/expensive and, thus, infeasible.}. Ultimately, running ML applications on edge devices can be significantly accelerated using such \textit{data-independent model compression} techniques~\cite{ddpaper} because the users trying to deploy a model on IoT devices will \textit{not} have to rely on the private datasets of third parties.\vspace{1mm}

\noindent
\textbf{Hardware constraints and network of devices in regards to inference:} As mentioned earlier, existing model compression techniques target efficient inference on a single device and not across a network of devices~\cite{deepComp, atkd, hintonKD}. This automatically opens up a new class of research problems -- \textit{communication-aware model compression}~\cite{nonnpaper}. For instance, modern smart home/cities applications can have many connected IoT sensors with, say, only 500KB total memory per node. To achieve high accuracy, the compressed deep networks often grow in size. Consequently, due to strict memory constraints, such models must be distributed across multiple nodes; this generates significant communication among the devices. Hence, the massive communication cost arising from this distributed inference presents a major (and so far largely ignored) impediment that prevents the effective utilization of the compute power of a network of IoT devices. Therefore, compressing models must not only account for hardware constraints, but also for the communication costs resulting from \textit{distributed inference}. In other words, since IoT consists of connected devices, this massive \textit{network} must be exploited to obtain true edge intelligence~\cite{bhardwaj2019edgeai}.

Fig.~\ref{flow} presents a unified view of the above challenges, as well as a few new research directions: (\textit{i})~FL-based training for preserving privacy while exploiting a network of devices (see Fig.~\ref{flow}(a), Section II), (\textit{ii})~Data-independent model compression for addressing hardware constraints without compromising data privacy (see Fig.~\ref{flow}(b), Section III), and (\textit{iii})~Communication-aware model compression to deploy deep learning models across a network of IoT devices (see Fig.~\ref{flow}(c), Section IV). Note that, existing computation-aware model compression techniques like~\cite{deepComp, atkd, hintonKD} only account for hardware constraints and do \textit{not} consider the other grand challenges in IoT era (see Fig.~\ref{flow}(d)). In this vision paper, we summarize our latest research while highlighting the importance of the above problems. We hope that these ideas will inspire future research in the field.

%
\section{Federated Learning (FL)}
\begin{figure}[]
\centering
\includegraphics[width=3.4in]{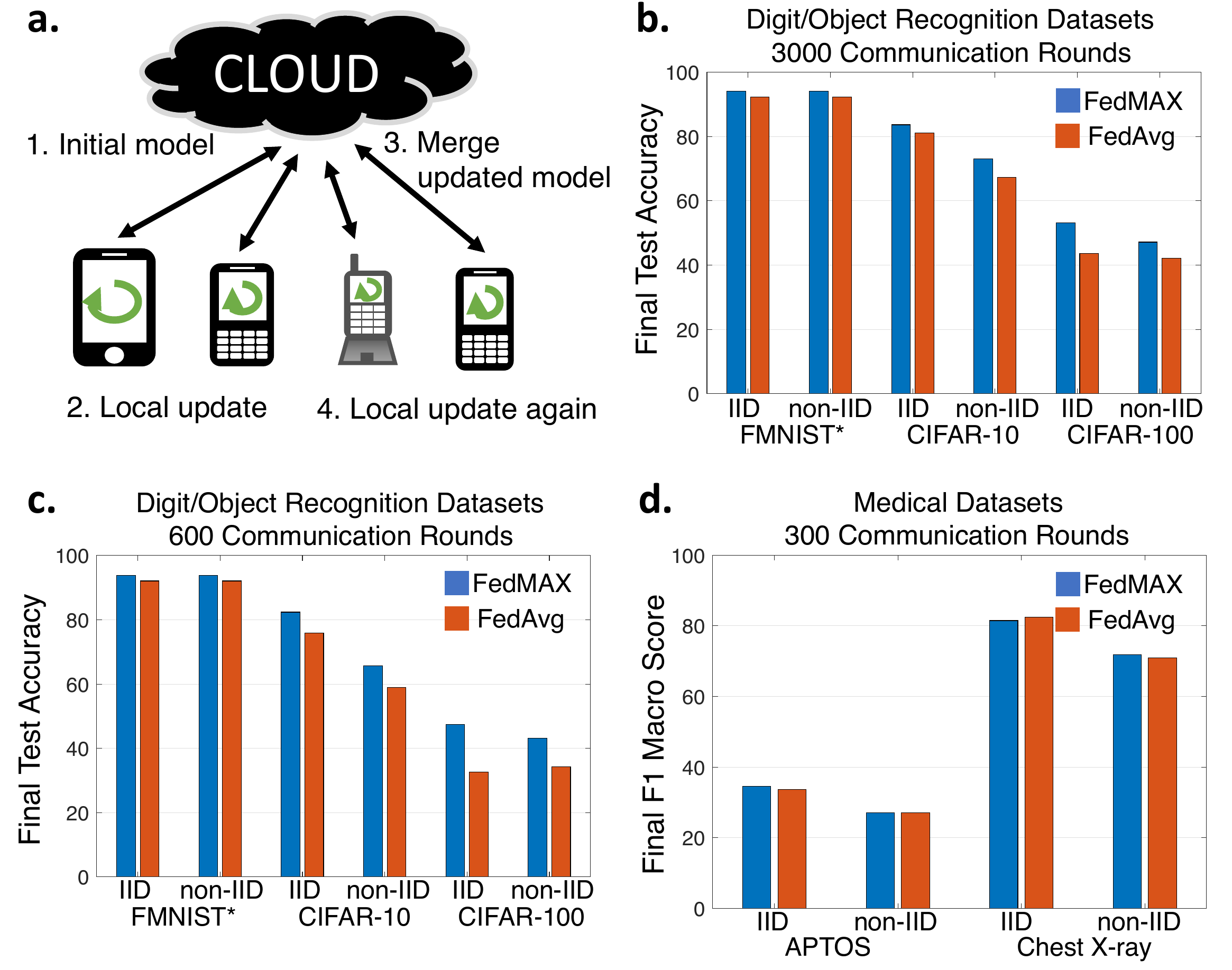}
	\caption{(a) Federated Learning: Several local training epochs are run on a randomly selected subset of devices. After local training, the models are sent to a server via a communication round; the server then averages the parameters of resulting local models to obtain a global model which is sent back to users. (b) Comparison between FedMAX and FedAvg for both IID and non-IID scenarios on three digit/object recognition datasets -- FEMNIST*, CIFAR-10 and CIFAR-100  -- for 3000 communication rounds. (c) Test accuracy for 600 communication rounds on the same datasets. (d) Test accuracy for different medical datasets, APTOS and Chest X-ray, with different approaches, FedAvg and our proposed approach (FedMAX) for 300 communication rounds.}
\label{fl}
\end{figure} 

Large amounts of data are increasingly generated on edge devices nowadays. However, since data on personal devices is highly sensitive, in order to preserve privacy, FL has become the de facto distributed training paradigm across a network of devices without sharing the data~\cite{konevcny2016federated,konevcny2015federated}. A typical FL framework is shown in Fig.~\ref{fl}(a). A model is trained on a random subset of local devices for some number of steps. Then, the model is sent to a server for a global update; in an established approach called FedAvg~\cite{konevcny2015federated}, the models from local devices are averaged at the server. Finally, this global model is sent back to the devices of individual users. 
We next explain a new technique called FedMAX that can be used to mitigate \textit{activation-divergence}, a new phenomenon which happens due to heterogeneity in data distributions~\cite{chen2020fedmax}.\vspace{2mm}

\noindent
\textbf{Activation-Divergence and FedMAX\vspace{1mm}} \\
\noindent
Since the data across users may not be independently and identically distributed (non-IID), local training updates can take the global model in different directions. As a result, when data is non-IID, the feature activation at the final layers of a model (\textit{e.g.}, at the output of fully-connected or average-pool layer\footnote{Average-pool layer in a CNN refers to the averaged output of final convolutional layer of the network (see Fig.~\ref{flow}(d)).}) can diverge across different users. Indeed, this makes the global model achieve lower accuracy. To address this activation-divergence issue, a prior based on the Principle of Maximum Entropy can be used~\cite{jaynes1957information}; this prior assumes minimal information about the local activations and aims to make activations for same classes similar across multiple devices. Since we exploit the Principle of Maximum Entropy, we call this approach FedMAX~\cite{chen2020fedmax}.

To demonstrate the effectiveness of FedMAX, we first show the test accuracy of both IID and non-IID scenarios on three different datasets: FEMNIST*~\cite{caldas2018leaf}, CIFAR-10 and CIFAR-100 in Fig.~\ref{fl}(b)(c). For each dataset, we train a CNN consisting of about 600K parameters. As evident, our approach significantly outperforms the prior FedAvg technique for all three datasets. Another important observation in Fig.~\ref{fl}(c) is that for all three datasets, FedMAX for 600 communication rounds achieves comparable or even better accuracy than FedAvg for 3000 communication rounds. Therefore, by relying on more localized training, FedMAX significantly reduces communication (by up to $5\times$) compared to prior techniques without losing accuracy.

A clear application for training on private datasets is in medical domain where the data is extremely sensitive. Hence, we also report experiments on two large medical datasets: APTOS (Retina images)~\cite{triastcyn2019federated} and Chest X-ray~\cite{kermany2018identifying}, which represent a real-life FL scenario. Since these are imbalanced datasets, we use the F1-macro score to measure the performance of the model. Also, pretrained ResNet50~\cite{he2016deep} models are fine tuned for these datasets.  
As shown in Fig.~\ref{fl}(d), the performance of FedMAX on the Chest X-ray dataset is close to FedAvg. One possible reason is that since the Chest X-ray dataset has only two classes, it can be unsuitable for making the activations more similar among labels across different clients. However, for the APTOS dataset, FedMAX outperforms FedAvg for both IID and non-IID settings. More experiments and details on FedMAX are given in~\cite{chen2020fedmax}. Therefore, by creating new, smarter loss functions, we can better exploit the network of devices for model training without compromising data privacy.

\section{Data-Independent Model Compression}
When deep learning models are trained on private datasets, the industries trying to deploy such models on edge devices cannot use the original datasets for model compression. Below, we describe our recent research targeting this problem\cite{ddpaper,bhardwaj2019edgeai}.\vspace{2mm}
\noindent
\textbf{Dream Distillation\vspace{1mm}} \\
\noindent
In Knowledge Distillation (KD)~\cite{hintonKD}, given a pretrained teacher model, a student model is trained using either the same dataset or using some unlabeled data (see Fig.~\ref{flow}(d)). For instance, to conduct KD on CIFAR-10 when real data is not available, \textit{alternate datasets} such as CIFAR-100 or tiny-Imagenet can be used to train a student model via KD~\cite{mismatch}. Prior work~\cite{hintonKD, mismatch} shows that the resulting student shows reasonably good accuracy on the original intended task, \textit{i.e.}, the CIFAR-10 test set classification. 
For many situations, however, it may not be possible to obtain even the alternate datasets. Then, \textit{how can we compress a deep network without using the original training set or any alternate data, without losing significant accuracy?} We answer this question in our recent work on \textit{Dream Distillation}~\cite{ddpaper}, a new technique which does not require real data for KD.

\begin{figure}[!t]
\centering
\includegraphics[width=3.4in]{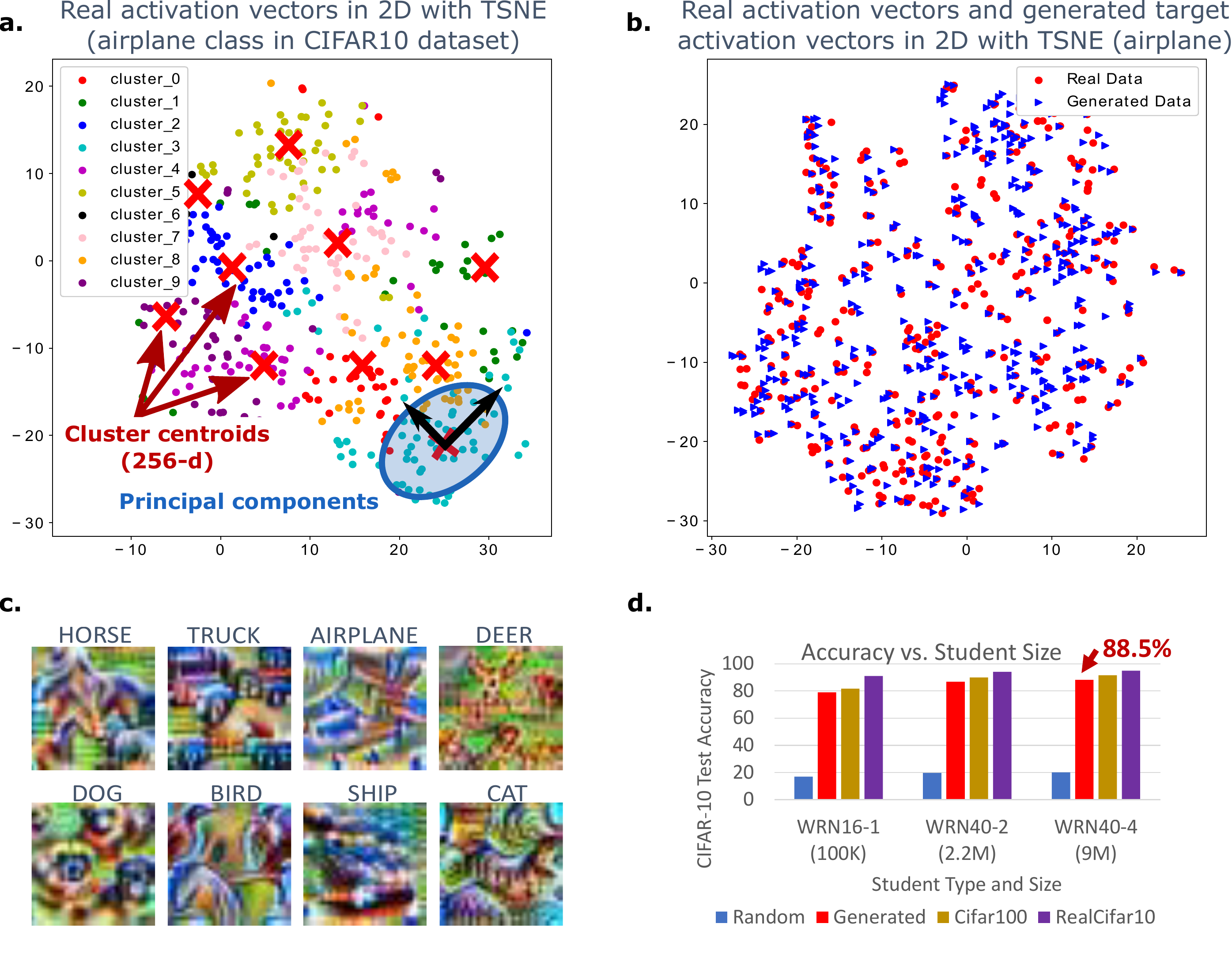}
	\caption{Dream Distillation: A data-independent knowledge distillation framework~\cite{ddpaper}. (a) Metadata used by our method, and (b) tSNE visualization of real data activations and target activations generated, (c) Examples of generated samples for CIFAR-10 dataset, and (d) Accuracy of student models trained from random data, generated data, alternate data: CIFAR100, and real CIFAR-10 dataset (Figure adopted from~\cite{bhardwaj2019edgeai}).}
\label{dd}
\end{figure}

Suppose our teacher network is a large Wide Resnet (WRN40-4, $8.9$M parameters), and the student model is a small Wide Resnet (WRN16-1, 100K parameters). The teacher network achieves $\sim95\%$ accuracy on CIFAR-10 test set. On the other hand, the student WRN16-1 model trained via Attention Transfer-based KD (ATKD)~\cite{atkd} with WRN40-4 teacher achieves about $91\%$ accuracy.  Most importantly, we assume that neither alternate data, nor original training set is available; rather, as explained below, only a small amount of metadata is given:
\begin{enumerate}
\item We use $k$-means algorithm to cluster the real activations at the average-pool layer of the teacher network for just $10\%$ of the real CIFAR-10 images. These cluster-centroids are used as metadata (see Fig.~\ref{dd}(a) for centroids of the airplane class in CIFAR-10).
\item Since the centroids represent the average activations in a cluster, they reduce the privacy-concerns as only the mean activations are used as metadata; the activations from real images are not used. Hence, this metadata does \textit{not} contain identifying information about the real images. 
\item Metadata also contains principal components for each cluster; this is illustrated in the Fig.~\ref{dd}(a) by the orthogonal principal components.
\end{enumerate}

\noindent
We use the teacher network and the above metadata to generate a large number of synthetic images which contain sufficient knowledge about the classes (even though they look far from real). Such images can be used to effectively distill knowledge from the teacher to the student without explicitly training the student network on any real data. 

Another prior work for data-independent model compression is Data-Free Knowledge Distillation (DFKD)~\cite{kd123} which also uses metadata. However, DFKD claims that using metadata from a single layer makes the problem under-constrained and, thus, leads to poor accuracy (\textit{e.g.}, DFKD achieves $68$-$77\%$ accuracy on MNIST). As a result, for more complex datasets like CIFAR-10, DFKD would achieve even lower accuracy. In contrast, our goal is to specifically demonstrate that metadata from a single layer is, in fact, sufficient to train accurate student models~\cite{ddpaper}.

Recently, deep learning visualization and interpretability~\cite{olah2018building} has received a lot of interest due to works like DeepDream~\cite{deepdream}. The objective of feature visualization is to generate an image that can maximize a certain objective (\textit{e.g.}, generate an image that can maximally activate a given hidden unit at a certain hidden layer, \textit{i.e.}, a neuron or a channel). 
DeepDream called these generated images as the \textit{Dreams} of the deep network. Since our approach exploits these ``dreams'' for KD, we call our approach Dream Distillation, as if we are distilling teacher's dreams!

In Dream Distillation, the first step is called \textit{Dream Generation} where we create custom objectives from the metadata. To this end, we add a small amount of Gaussian noise to the cluster-centroids along the principal component directions to create target activations. The resulting distributions of target activations and real activations are very similar (see Fig.~\ref{dd}(b)). Next, we generate $50,000$ images such that the average-pool activations of generated images layer are as close as possible to the target activations. Finally, these synthetic images are used for KD between teacher and student networks. More details on the approach are given in~\cite{ddpaper}.

Some examples of synthetic images created by our Dream Generation technique are shown in Fig.~\ref{dd}(c). We note that for classes like \{horse, truck, deer, dog\}, some key features (\textit{e.g.}, animal faces, wheels, \textit{etc.}) are clearly visible. However, for classes like \{ship, airplane, bird, cat\}, the images are significantly more subtle. For instance, the teacher network mostly generates a striped pattern for cats (instead of more obvious features such as cat faces, \textit{etc.}). Given how different these synthetic images are from natural images, we next investigate if these images be used for distilling knowledge from teacher to the student.

To evaluate the effectiveness of Dream Distillation, we conduct KD using four datasets containing $50,000$ images: (\textit{i}) Random noise images, (\textit{ii}) Dream Distillation images, (\textit{iii}) CIFAR-100 images as an alternate dataset, and (\textit{iv}) Real CIFAR-10 dataset. We use three student models: WRN16-1 (100K parameters), WRN40-2 (2.2M parameters), and WRN40-4 (8.9M parameters). For the last case, both the teacher model and the student model are the same. 
Fig.~\ref{dd}(d) shows the test accuracy for all student models. Clearly, Dream Distillation performs comparable to the alternate dataset CIFAR-100. 
More specifically, using synthetic images generated by our model,  WRN16-1 student achieves $\sim79\%$ accuracy, while using CIFAR-100 images, it achieves $\sim81\%$ accuracy on CIFAR-10 test set. Interestingly, our WRN40-4 student trained via Dream Distillation achieves $88.5\%$ accuracy on CIFAR-10 test set without ever seeing \textit{any} real data!
%
Hence, the synthetic images generated via our method can transfer significant amount of relevant knowledge about the real data without requiring any real or alternate datasets. Therefore, Dream Distillation can allow industries to more rapidly compress and deploy deep learning models without access to third-party proprietary data.






\section{Communication-Aware Model Compression}
In the IoT era, the network of edge devices must be exploited via distributed learning to obtain high-accuracy, low-latency intelligence. Most of the model compression literature focuses on computational aspects such as energy, memory, and latency of inference on a \textit{single} device~\cite{deepComp, hintonKD}. As a result of this lack of network-aware algorithms, the prior art for communication-aware deployment of deep learning models (across \textit{multiple} devices) is significantly limited. 
For instance, a recent distributed inference method called SplitNet aims to split a deep learning model into disjoint subsets without considering the strict memory- and FLOP-budgets for IoT devices~\cite{splitnet}. Consequently, the disjoint models obtained by SplitNet may not satisfy the strict memory constraints of edge devices. Similarly, MoDNN~\cite{modnn} aims to reduce the number of FLOPS during distributed inference. However, MoDNN assumes that the entire model can fit on each device, and does not consider any model compression. 

The assumption that the entire model can fit on each IoT-device is optimistic because many IoT devices are significantly memory-constrained. In fact, in such memory-constrained scenarios, the model itself must be distributed across multiple devices which can lead to heavy inter-device communication at each layer of the deep network. Therefore, a new paradigm is needed to not only reduce memory and computation of deep networks, but also to minimize communication for efficient distributed inference. In other words, the network of IoT devices must be exploited to improve the accuracy without increasing the communication latency. Towards this communication-aware model compression, we present our recent idea on Network-of-Neural Networks (NoNN)~\cite{nonnpaper, bhardwaj2019edgeai}.\vspace{2mm}

\noindent
\textbf{Network-of-Neural Networks (NoNN)\vspace{1mm}} \\
\noindent
NoNN refers to a new distributed inference paradigm which enables new memory- and communication-aware student architectures obtained from a single large teacher model. Specifically, a NoNN consists of a collection of multiple, disjoint student modules which focus only on a part of teacher's knowledge. 
Individual students are deployed on separate edge devices to \textit{collaboratively} perform the distributed inference. 

\begin{figure}[]
\centering
\includegraphics[width=3.4in]{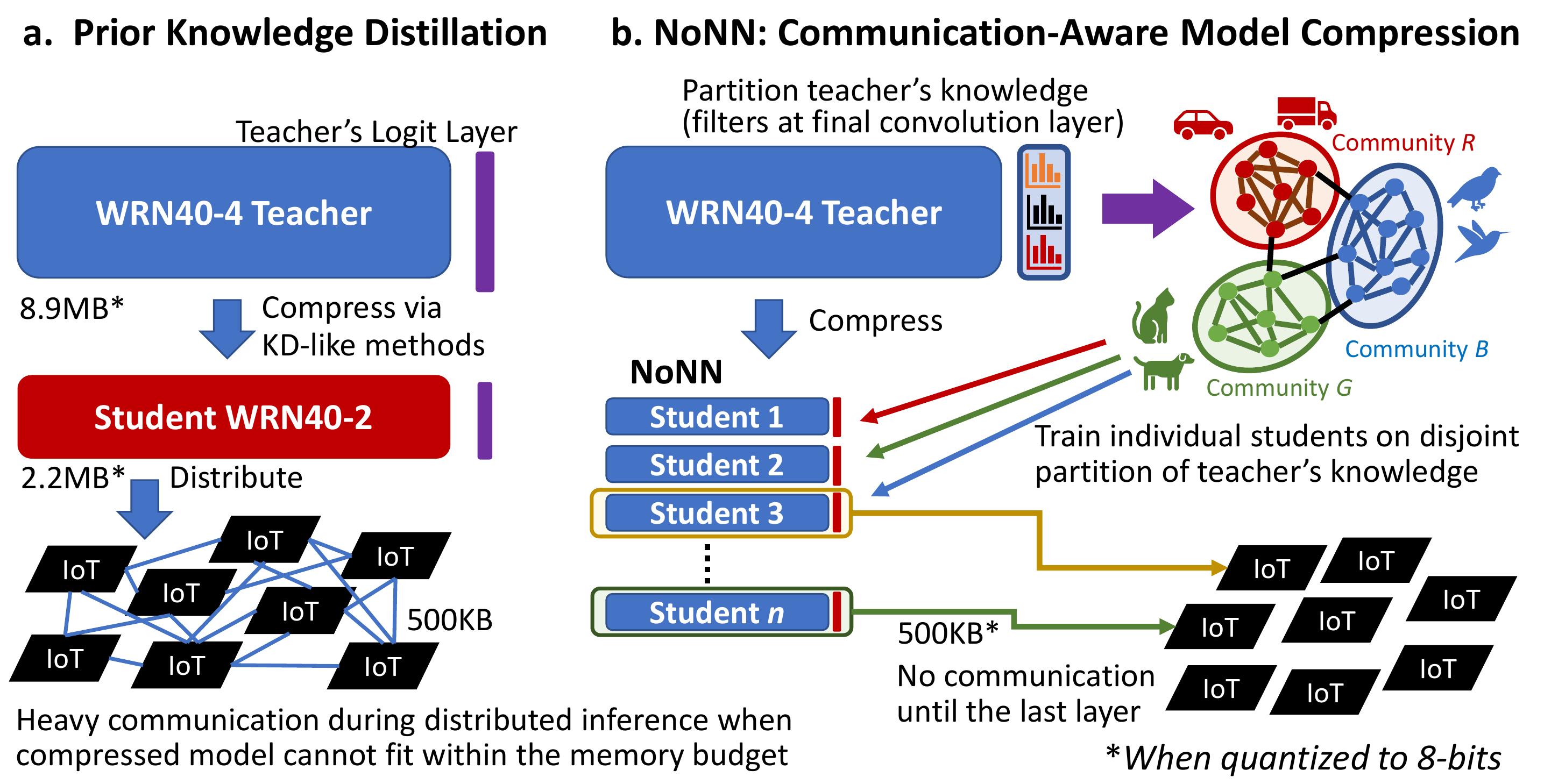}\vspace{-2mm}
	\caption{(a) Prior KD~\cite{hintonKD}: Distributing large student models that do not fit on a single memory-limited IoT device leads to heavy communication at each layer. (b) NoNN~\cite{nonnpaper} results in disjoint students that can fit on individual devices: Filters at teacher's final convolution layer (representing knowledge about different classes) can be partitioned to train individual students which results in minimal communication until the final layer (Figure adopted from~\cite{bhardwaj2019edgeai}).\vspace{-2mm}}
\label{nonn}
\end{figure}

Key differences between traditional KD and NoNN are shown in Fig.~\ref{nonn}. As evident, NoNN relies on concepts from \textit{network science}~\cite{networksci} to partition teacher's final convolution layer. Specifically, features for various classes are learned at different filters in CNNs. These activation patterns reveal how teacher's knowledge gets distributed at the final convolution layer. Therefore, such activation patterns can be used to create a \textit{filter activation network}~\cite{nonnpaper} that represents how the teacher's knowledge about various classes is organized into filters (see Fig.~\ref{nonn}(b)). We then partition this network via community detection~\cite{networksci} (more details are given in~\cite{nonnpaper}); these disjoint partitions of teacher's knowledge are used to train individual student modules.

Since we train completely separate students to mimic parts of teacher's knowledge, our NoNN results in a highly parallel student architecture as shown in Fig.~\ref{nonn}(b). Moreover, we can select significantly smaller individual student modules (subject to some memory/FLOP budgets). Consequently, the NoNN incurs significantly lower memory, computations, and communication. In other words, our individual student modules adhere to the memory- and FLOP-constraints of IoT devices, and do not communicate until the final fully connected layer.

To better illustrate these ideas, we now discuss a proof-of-concept for NoNN using the CIFAR-10 dataset. Assuming each IoT-device has a memory budget of 500KB, each student module in NoNN must have less than 500K parameters\footnote{CNNs with 500K parameters can fit within 500KB with 8-bit quantization.}. Again, we choose our teacher model as Wide Resnet WRN40-4 (8.9M parameters), and the NoNN consists of two student modules, each based on a Wide Resnet structure (denoted as NoNN-2S)~\cite{nonnpaper}. Our NoNN-2S model has  $\sim860$K total parameters, where each student module has $\sim430$K parameters. We further use TVM\footnote{TVM compiler: \url{https://tvm.ai/}} to deploy all models on Raspberry-Pi (RPi) devices in order to demonstrate the effectiveness of NoNN for distributed inference. We connect RPi's via a point-to-point wired connection to represent a network of edge devices. 

The results for the teacher model deployed on one RPi and our NoNN-2S model deployed on two RPi's are shown in Table~\ref{nt} (top). The NoNN-2S model clearly outperforms the teacher network by a very large margin ($13\times$ to $20\times$ improvement) in memory, FLOPS, and energy with only $1\%$ loss of accuracy. Overall, our experiments show great agreement between theory and practice. Specifically, while the theoretical improvement (per student) in FLOPS is about $15\times$, we achieved $14\times$ better energy consumption in practice. We thoroughly demonstrate the effectiveness of NoNN on five well-known image classification tasks and also show results for a higher number of students in~\cite{nonnpaper}. A hardware prototype of NoNN is further presented in~\cite{nonnDemo}. 
\begin{table}[]
\centering
\caption{Communication-Aware Model Compression: CIFAR-10 Results~\cite{nonnpaper} (Table adopted from~\cite{bhardwaj2019edgeai})\vspace{-2mm}}
\scalebox{0.897}{
\label{nt}
\begin{tabular}{|l|c|c|c|} 
\hline
\multirow{2}{*}{TVM Setup}& WRN40-4 Teacher & NoNN-2S (one student & \multirow{2}{*}{Gain}\\
& (on a single RPi)~ & on each RPi)$^*$& \\
\hline \hline
$\#$parameters& $8.9$M& $\bm{0.43}$\textbf{M}& $\bm{20.7\times}$\\ \hline
$\#$FLOPS& $2.6$G& $\bm{167}$\textbf{M}& $\bm{15.5\times}$\\ \hline
Latency (ms)& $1405$& $\bm{115}$& $\bm{12.2\times}$\\ \hline
Energy (mJ)& $3430.67$& $\bm{238.98}$& $\bm{14.3\times}$\\ \hline
Accuracy & $\bm{95.49\%}$& $94.32\%$& $-1.17\%$\\ \hline \hline
\end{tabular}
}
\scalebox{0.896}{
\begin{tabular}{|l|c|c||c|}
\multirow{2}{*}{Pytorch Setup} & \multicolumn{2}{c||}{~Split-ATKD (WRN40-2)} & ~NoNN-8S~ \\ \cline{2-3} 
 & 4 RPi's & 8 RPi's & 8 RPi's\\ \hline \hline
Accuracy & $95.03\%$ & $95.03\%$ & $95.02\%$ \\ \hline
Parameters per device & $550$K & $275$K & $430$K \\ \hline
FLOPS per device & $163$M & $82$M & $167$M \\ \hline
Total latency per inference (s)~ & $23$ & $28.5$ & $\bm{0.85}$ \\ \hline \hline
Speedup with NoNN & $\bm{27\times}$ & $\bm{33\times}$ & $-$ \\ \hline
\end{tabular}
}
\begin{flushleft}
{\scriptsize $^{*}$These results are for each individual student. For complete NoNN, $\#$parameters will be $\sim0.8$M, and FLOPS and total energy (for both RPi's) will double. However, this is acceptable since we are concerned with per-device memory and compute budgets.\vspace{-2mm}}
\end{flushleft}
\end{table}

Finally, we note that a horizontally-split deep network will lead to heavy communication cost. 
To demonstrate this, we train and deploy a NoNN-8S model across a network of eight RPi's. We then distribute an ATKD-based compressed model (WRN40-2, 2.2M parameters, $95.03\%$ accuracy)~\cite{atkd} on four and eight RPi's for comparison\footnote{All splitting experiments are performed in Pytorch (including for NoNN-8S) because TVM binary cannot be distributed across multiple devices~\cite{nonnpaper}.}. Table~\ref{nt} (bottom) summarizes the results of distributed inference using NoNN-8S and split-ATKD. As evident, while achieving a similar accuracy of $95.02\%$, NoNN-8S is $27$-$33\times$ faster than the split-ATKD models, even though FLOPS per device for split-ATKD are significantly lower. Therefore, when the compressed models cannot fit within the memory budget of IoT devices, communication can indeed significantly degrade the inference latency. This clearly underscores the importance and superiority of our communication-aware model compression.

\section{Conclusion}
In this paper, we have highlighted three major challenges to large-scale adoption of deep learning at the edge: (\textit{i})~Hardware constraints of IoT devices, (\textit{ii})~Data privacy in the IoT era, and (\textit{iii})~The lack of network-aware deep learning algorithms as most inference techniques focus on single devices. Since these challenges are closely intertwined with deep learning training and inference, we have then provided a unified view of emerging research directions occurring at their intersection. Specifically, we have discussed: (1)~FL for training deep networks, (2)~Data-independent deployment of learning, and (3)~Communication-aware distributed inference. Ultimately, our vision aims to bring the network at the forefront of IoT in order to truly enable intelligence at the edge.

%

\ifCLASSOPTIONcompsoc
  \section*{Acknowledgments}
  We thank the anonymous reviewers for their useful suggestions which helped improve the paper. We also acknowledge Amazon support through the AWS ML Research Program.
\else
  \section*{Acknowledgment}
  This material is based on research sponsored in part by the Air Force Research Laboratory (AFRL) and Defense Advanced Research Projects Agency (DARPA) under agreement number FA8650-18-2-7860. The U.S. Government is authorized to reproduce and distribute reprints for Governmental purposes notwithstanding any copyright notation thereon. The views and conclusions contained herein are those of the authors and should not be interpreted as necessarily representing the official policies or endorsements, either expressed or implied, of Air Force Research Laboratory (AFRL) and Defense Advanced Research Projects Agency (DARPA) or the U.S. Government. We also thank NVIDIA Corporation for donating a Titan Xp GPU which was used for conducting some experiments. Finally, we acknowledge Amazon support through the AWS MLRA program.\vspace{0.5mm}
\fi


\ifCLASSOPTIONcaptionsoff
  \newpage
\fi



%

\bibliographystyle{abbrv}
\bibliography{sample-bibliography}

%
%

%

\end{document}